\documentclass[runningheads]{llncs}
\usepackage[T1]{fontenc}
\usepackage{graphicx,verbatim}
\usepackage{hyperref}
\usepackage{bbding}
\usepackage{amsmath}
\usepackage{amsfonts}
\usepackage{array}
\usepackage{xcolor}
\usepackage{multirow}
\usepackage{booktabs}
\usepackage{cite}
\usepackage{amssymb}

\usepackage{marvosym}

\begin{document}
\title{GraphMMP: A Graph Neural Network Model with Mutual Information and Global Fusion for Multimodal Medical Prognosis}
\titlerunning{GraphMMP: A GNN Model for Multimodal Prognosis}

\author{Xuhao Shan\inst{1} \and
Ruiquan Ge\inst{1,}\textsuperscript{\Letter} \and
Jikui Liu\inst{2} \and
Linglong Wu\inst{1} \and
Chi Zhang\inst{3} \and
Siqi Liu\inst{4} \and
Wenjian Qin\inst{5} \and
Wenwen Min\inst{6} \and
Ahmed Elazab\inst{7} \and
Changmiao Wang\inst{4,}\textsuperscript{\Letter}}

\authorrunning{X. Shan et al.}

\institute{Hangzhou Dianzi University, Hangzhou, China \\
\email{gespring@hdu.edu.cn}
\and
Shenzhen Polytechnic University, Shenzhen, China \and
The Chinese University of Hong Kong, Shenzhen, China \and
Shenzhen Research Institute of Big Data, Shenzhen, China \\
\email{cmwangalbert@gmail.com} \and
Shenzhen Institutes of Advanced Technology, Shenzhen, China \and
Yunnan University, Kunming, China \and
Shenzhen University, Shenzhen, China
}

\maketitle              % typeset the header of the contribution
\begin{abstract}
In the field of multimodal medical data analysis, leveraging diverse types of data and understanding their hidden relationships continues to be a research focus. The main challenges lie in effectively modeling the complex interactions between heterogeneous data modalities with distinct characteristics while capturing both local and global dependencies across modalities. To address these challenges, this paper presents a two-stage multimodal prognosis model, GraphMMP, which is based on graph neural networks. The proposed model constructs feature graphs using mutual information and features a global fusion module built on Mamba, which significantly boosts prognosis performance. Empirical results show that GraphMMP surpasses existing methods on datasets related to liver prognosis and the METABRIC study, demonstrating its effectiveness in multimodal medical prognosis tasks. Our code is publicly available on \url{https://github.com/Windbelll/GraphMMP}.

% Authors must provide keywords and are not allowed to remove this Keyword section.

\end{abstract}
\section{Introduction}

Deep learning has emerged as a powerful tool for addressing complex medical challenges \cite{Topol2023AsAI}. Early methods typically rely on single-modality methods, such as ResNet50~\cite{he2016deep} and Vision Transformer (ViT)~\cite{dosovitskiy2020image} for image analysis, or Support Vector Machine (SVM)~\cite{Cortes1995SupportVectorN} and Random Forest (RF)~\cite{Breiman2001RandomF} for radiomics data processing. Unlike traditional single-modality approaches, deep learning excels at integrating and analyzing heterogeneous data, including medical imaging, genomic sequences, clinical diagnoses, and electronic health records \cite{Behrad2022AnOO}. This ability to extract relevant features has made multimodal medical data analysis a central focus in medical artificial intelligence.

In the medical field, researchers have proposed various data fusion techniques, such as SimpleFF~\cite{choi2021fully} and HFBSurv~\cite{li2022hfbsurv}, which fuse different modalities by combining deep learning and radiomics or using factorized bilinear models. Methods such as MMD~\cite{cui2022survival} and TMI-CLNet~\cite{wu2025tmi} improve modality fusion by introducing cross-attention mechanisms. However, traditional convolutional neural network methods are usually unable to effectively solve these problems because they have limitations in processing complex and heterogeneous data, especially when fusing multimodal data, and cannot capture the global relationship between modalities. Graph Neural Networks (GNNs) possess a unique ability to effectively process intrinsic connections within data, thanks to their inherent graph structure, such as GraphConv~\cite{Morris2018WeisfeilerAL} and UniMP~\cite{shi2021masked}, have enhanced model performance by leveraging graph convolutional operations and incorporating masked labels, respectively~\cite{Velickovic2023EverythingIC}. 

Furthermore, the application of GNNs in the medical field has gradually expanded. For example, PrimeKG supports AI-based drug-disease interaction analysis by integrating multiple biomedical resources to provide support for personalized treatment~\cite{Chandak2022BuildingAK}. Meanwhile, CGMega improves cancer gene module prediction by combining multi-omics data with graph attention mechanisms~\cite{Li2024CGMegaEG}, while BrainGNN uses GNNs to identify brain regions related to the nervous system in the field of neurological diseases~\cite{li2021braingnn}. Despite these methods demonstrating the great potential of GNNs in medical research, they usually rely on simple feature map construction methods, which cannot fully capture the implicit information between modalities and have deficiencies in modeling global relationships. Specifically, the limited perception of global information across different modalities prevents the model from effectively leveraging global features for optimization.

To address the aforementioned limitations, we propose a novel two-stage medical multimodal prognosis model based on GNNs, motivated by their inherent capability to effectively integrate heterogeneous data. Our key contributions are: (1) We introduce a two-stage medical multimodal prognosis model that effectively tackles challenges in heterogeneous data fusion. (2) We develop a feature graph construction mechanism based on mutual information. This mechanism enables the feature map to capture hidden inter-modal information, significantly improving prognosis performance. (3) We design a global fusion module based on Mamba \cite{gu2023mamba}, which enhances the GNN's global perception capability and boosts the overall performance.

\section{Methods}

\begin{figure}
\includegraphics[width=1\textwidth]{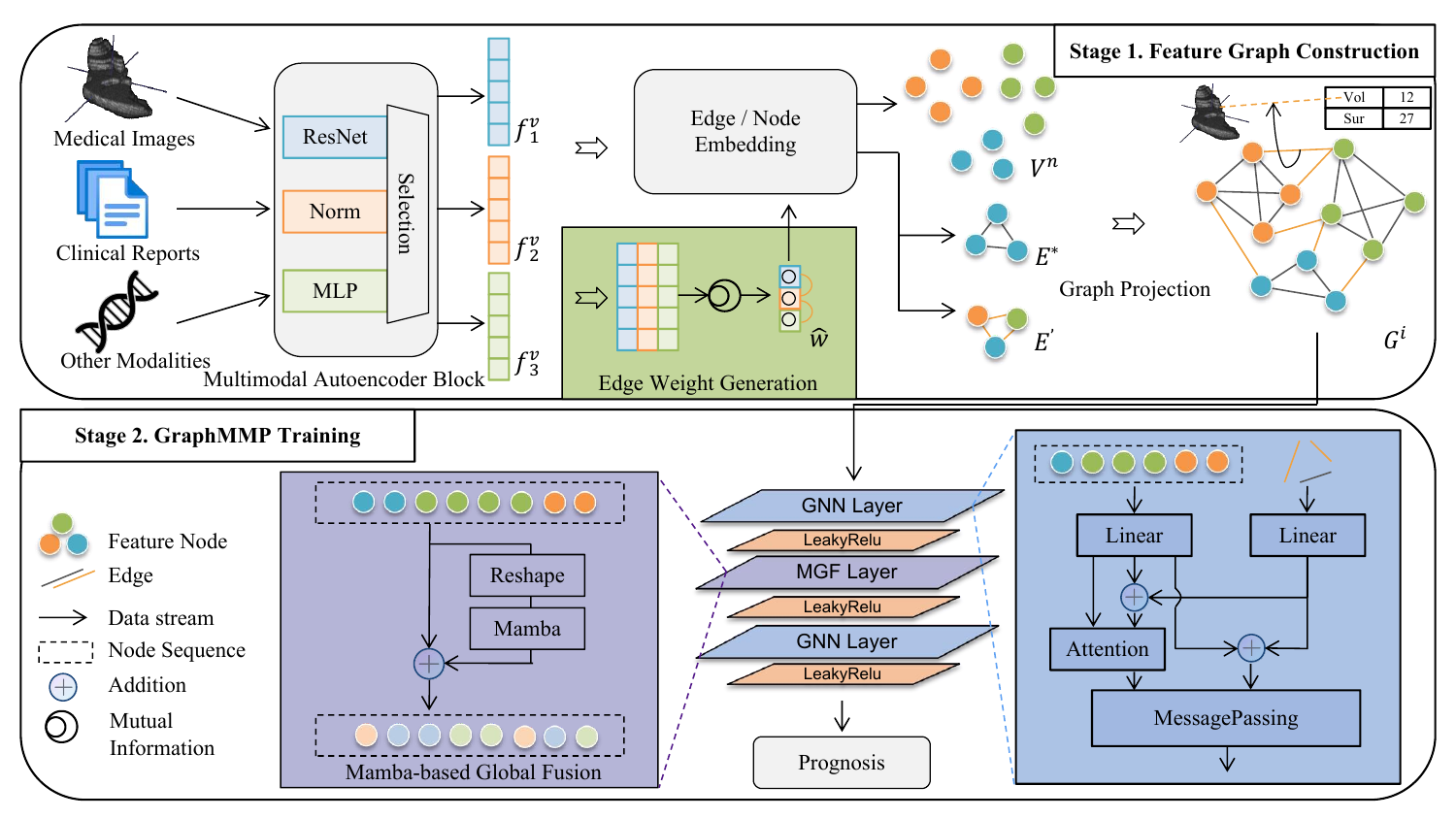}
\caption{Overall framework of the proposed GraphMMP. The framework includes two stages: (1) constructing a feature graph from multimodal data using mutual information, and (2) training GraphMMP with a Mamba-based global fusion module for prediction.}% The input multimodal features are initially processed into a feature graph during the first stage of graph construction, which is suinalbsequently fed into the main GraphMMP network in the second stage for training, culminating in the derivation of predictive outcomes.} 
\label{arch}
\end{figure}

This paper proposes a two-stage Graph-based Multimodal Medical Prognosis model (GraphMMP). The architecture of the proposed GraphMMP, depicted in Fig.~\ref{arch}, consists of two stages: Stage 1, Feature Graph Construction, and Stage 2, GraphMMP Training. Let \( \mathcal{P} = \{\mathbf{P}_1, \mathbf{P}_2, \ldots, \mathbf{P}_N\} \) represents a dataset containing patient samples, where \( N \) is the total number of subjects. Each patient sample is defined as \( \mathbf{P}_i = \{\mathbf{X}^1_i, \mathbf{X}^2_i, \mathbf{X}^3_i, \ldots, \mathbf{X}^K_i, \mathbf{Y}_i\} \), with \( K \) denoting the total number of modalities. Here, \( \mathbf{X}^k_i \) corresponds to multimodal prognostic factors, and \( \mathbf{Y}_i \) represents the prognostic outcome. In Stage 1, we convert the raw data \( \mathbf{P}_i \) into feature graphs \( \mathbf{G}_i \), which encapsulate high-dimensional semantic representations. In Stage 2, we design a GNN \( \mathcal{H} \) to ensure the model’s predictions align with the actual prognostic outcomes. This objective is formalized as optimizing the following mapping:
\begin{equation}
\mathbf{Y}_i = \mathcal{H}(\mathbf{G}_i),
\end{equation}
where the network \( \mathcal{H} \) learns to project the feature graphs \( \mathbf{G}_i \) onto their corresponding prognostic outcomes \( \mathbf{Y}_i \). 
\subsection{Feature Graph Construction}
\label{sec2.1}

At this stage, the multimodal data undergoes initial processing through the Multimodal Autoencoder Block (MAB) for feature modeling and extraction. This step produces feature vectors that capture high-dimensional semantic information from the raw data. The MAB performs modality-specific feature extraction using pre-trained encoders, which remain fixed and are not involved in further learning. For instance, image data is processed using a ResNet architecture followed by flattening, while clinical data metrics are standardized to create their respective feature vectors \( \mathbf{f}^v_i \), where \( i \) represents the index of different modalities. This feature extraction process aims to reduce the dimensionality of the feature vectors, making it easier to construct subsequent feature graphs on a more manageable scale. After this step, node embeddings and edge embeddings are generated based on these feature vectors.

\noindent\textbf{Node Embeddings.}  
During the node embedding process, these feature vectors are combined into a node feature matrix \(\mathcal{X} \in \mathbb{R}^{M \times 1}\) using the following formula:

\begin{equation}
\mathcal{X} = \bigoplus_{i=1}^{K} \mathbf{f}^v_i,
\end{equation}

\noindent where \(M = \sum \text{len}(\mathbf{f}^v_i)\) represents the total number of nodes, and \(\bigoplus\) signifies the concatenation of features.

\noindent\textbf{Edge Weight Generation.}  
The strength of associations between features is quantified using Mutual Information (MI), which measures the dependency between modality feature vectors. The MI \( I(\mathbf{f}^v_i; \mathbf{f}^v_j) \) between two feature vectors \( \mathbf{f}^v_i \) and \( \mathbf{f}^v_j \) is calculated as follows:

\begin{equation}
I(\mathbf{f}^v_i; \mathbf{f}^v_j) = \sum_{x \in \mathbf{f}^v_i} \sum_{y \in \mathbf{f}^v_j} p(x, y) \log \frac{p(x, y)}{p(x)p(y)},
\end{equation}

\noindent where \( p(x, y) \) represents the joint probability distribution of \( \mathbf{f}^v_i \) and \( \mathbf{f}^v_j \), and \( p(x) \) and \( p(y) \) denote their respective marginal probability distributions. Once the MI values are computed, the feature weights, representing the relationships between and within modalities, are computed using the following formula:

\begin{equation}
w_{i,j} = \sigma\left(I(\mathbf{f}^v_i; \mathbf{f}^v_j)\right),
\end{equation}

\noindent where \(\sigma(x) = \frac{1}{1 + e^{-x}}\) is the sigmoid function. To ensure numerical stability and prevent excessive small weights, a threshold \(\tau = 0.1\) is used via weight clipping:

\begin{equation}
\hat{w} = \max(w, \tau).
\end{equation}

This method incorporates implicit inter-modal feature relationships into the feature graph through MI. By doing so, it enhances the graph's ability to represent and propagate information effectively between nodes from different modalities, thereby improving its overall expressive power.

\noindent\textbf{Edge Embeddings.}  
To create edge embeddings, we strategically connect nodes to manage the feature graph's complexity. Nodes belonging to the same modality are fully connected, represented as \(\mathbf{E}^*\). Meanwhile, sparse connections are probabilistically established between nodes from different modalities, denoted as \(\mathbf{E}'\). The weighted edges, \(\mathcal{E}\), are thus defined as:

\begin{equation}
\mathcal{E} = (\mathbf{E}^* \cup \mathbf{E}', \hat{w}).
\end{equation}

With these connections in place, the feature graph \(\mathcal{G} = (\mathcal{X}, \mathcal{E})\) is constructed, integrating both the node features and their relationships.

\subsection{GraphMMP Training}

After feature graph construction, we develop a lightweight GNN, GraphMMP, specifically designed for medical prognosis tasks. The GraphMMP architecture comprises two GNN layers followed by a Mamba-based Global Fusion (MGF) layer. To enhance gradient flow and improve overall performance, we use LeakyReLU activation functions between these layers. The network concludes with a classification head, which generates the final classification results.

\noindent\textbf{GNN Layer.} GNNs excel at processing multimodal feature graphs by gathering information from adjacent nodes through message-passing, capturing complex relationships within multimodal data. Inspired by Transformer Graph \cite{Hu2020HeterogeneousGT}, we introduce a multi-head attention mechanism that computes attention weights for neighboring features, where both edge weights and neighboring nodes contribute to message aggregation, enhancing the graph representation. This attention is represented using:

\begin{equation}
\alpha_{ij} = \text{Softmax}\left( \frac{Q_i \cdot (K_j + E_{ij})^T}{\sqrt{d}} \right),
\end{equation}
where \(\alpha_{ij}\) denotes the attention weight between node \(i\) and its neighboring node \(j\), \(Q_i\) and \(K_j\) represent the Query and Key of these nodes, respectively, and \(E_{ij}\) is the edge weight connecting them. In the message aggregation process, the value \(V_j\) of each neighboring node is combined with the edge weight \(E_{ij}\) and scaled by the attention weight \(\alpha_{ij}\). This updated information contributes to the new representation of node \(i\), as formalized by:

\begin{equation}
h'_i = \left( \sum_{j \in \mathcal{N}(i)} \alpha_{ij} (V_j + E_{ij}) \right) \, \| \, h_i,
\end{equation}
where \(\|\) signifies the concatenation operation and \(h'_i\) is the updated representation of node \(i\).

\noindent\textbf{Mamba-based Global Fusion.} Despite their strengths, GNNs struggle to capture deep inter-modal correlations from a global perspective, limiting model performance. To address this, we introduce the MGF module, which leverages Mamba, a novel sequence model designed to focus on task-relevant information while ignoring irrelevant details \cite{gu2023mamba}. The MGF module processes the node features by first flattening the feature graph nodes into a sequential format, which is then passed through the Mamba block to extract global feature information. This global information is combined with the original node sequence via a residual connection, and the enhanced sequence is reshaped back into the feature graph structure. This process enriches the nodes with global contextual features, improving the model's ability to capture inter-modal correlations and enhancing overall performance.

\section{Experiments}
\subsection{Dataset and Implementation}

Our study utilizes two datasets, both of which were preprocessed and converted into a unified graph data format to ensure consistency across experiments.

\textbf{Liver Prognosis Dataset.}  
The first dataset is a private liver prognosis dataset from a partner hospital. After obtaining informed consent, the data were collected anonymously, with all personally identifiable information removed. The dataset covers a range of liver disease etiologies. After initial screening and data cleaning, the final dataset consists of 185 patient cases. Each case includes CT images, radiomics quantitative data, and clinical information such as biochemical analyses, fat analysis, and demographic details, a total of five modalities.

\textbf{METABRIC Dataset \cite{curtis2012genomic}.}  
The second dataset is the Molecular Taxonomy of Breast Cancer International Consortium (METABRIC), a widely recognized resource in breast cancer research. METABRIC provides extensive clinical, pathological, and molecular data, aiming to elucidate the molecular features of breast cancer and their connection with patient outcomes. For this dataset, we used gene expression data, copy number variation data, and clinical data. The 5-year prognosis outcome was selected as the target label for our analysis.

\textbf{Implementation Details. }All experiments were performed on a consistent platform, utilizing an Ubuntu 20.04 server operating system and an NVIDIA 3090Ti GPU. The software environment included CUDA version 11.8, PyTorch version 2.0, and PyTorch Geometric version 2.6.1. All experiments were conducted using 5-fold cross-validation to ensure the model's stability and generalization capabilities. The learning rate was set to \(1 \times 10^{-4}\), and the model was trained for 200 epochs. For parameter optimization, we used the Adam optimizer, and cross-entropy served as the loss function. To prevent overfitting, an early stopping mechanism with a patience of 50 epochs was implemented. For performance evaluation, we used Accuracy (ACC), Precision, Recall, F1-score, and Area Under the Curve (AUC).

\subsection{Experimental Results}

\begin{table}[t] 
\centering
\caption{Comparisons with other methods using various modality configurations on our private dataset. I and T represent image data and non-image information, respectively. The best and second-best results are shown in \textcolor{red}{red} and \textcolor{blue}{blue}, respectively.}
\begin{tabular}{ccccccc}
\hline
Method & Modality & ACC & Precision & Recall & F1-score& AUC\\ 
\hline
GraphConv~\cite{Morris2018WeisfeilerAL}	&T	&0.7823
& 0.8252 & 0.5883	&0.6708	&0.7661\\
UniMP~\cite{shi2021masked}	&T	&0.7931& 0.7962 & 0.6797 &0.7147	&0.7861\\ 
GraphMMP (Ours) & T &0.8099& 0.8182 & 0.6429&0.7222 &0.8037 \\
\hline
SimpleFF~\cite{choi2021fully} & I+T &0.7275&0.7472&	0.5410&	0.6218&	0.7050\\ 
HFBSurv~\cite{li2022hfbsurv}	 &I+T	&0.7945	&0.8201	&0.6432	&0.7233	&0.8015\\
MMD~\cite{cui2022survival}	& I+T	&0.7722	&0.8125	&0.6063&0.6772	& 0.7344 \\ 
TMI-CLNet~\cite{wu2025tmi} & I+T &\textcolor{blue}{0.8312}	&\textcolor{blue}{0.8438} &\textcolor{blue}{0.7428} &\textcolor{blue}{0.7805}&\textcolor{blue}{0.8223} \\ 
GraphMMP (Ours) & I+T &\textcolor{red}{0.8514} & \textcolor{red}{0.8462} & \textcolor{red}{0.7857}&\textcolor{red}{0.8106}&\textcolor{red}{0.8411} \\ \hline
\end{tabular}
\label{comparison}
\end{table}

\begin{table}[t] 
\centering
\caption{Comparison with other graph-based methods on the METABRIC dataset. The best and second-best results are shown in \textcolor{red}{red} and \textcolor{blue}{blue}, respectively.}
\begin{tabular}{cccccc}
\hline
Method& ACC & Precision & Recall & F1-score& AUC\\ 
\hline
GraphConv~\cite{Morris2018WeisfeilerAL}&0.8056&0.6571&0.4646&0.5562&0.8502\\
UniMP~\cite{shi2021masked}&\textcolor{blue}{0.8369}&\textcolor{blue}{0.7263}&\textcolor{blue}{0.6346}&\textcolor{blue}{0.6819}&\textcolor{blue}{0.8573}\\
GCN~\cite{Kipf2016SemiSupervisedCW} &0.7626&0.6111&0.1112&0.1897&0.7116\\ 
ChoqFuzGCN~\cite{Palmal2023BreastCS} &0.8200&0.7142&0.5983&0.6470&0.8301\\ 
MOGAT~\cite{Tanvir2024MOGATAM} &0.8056&0.6279&0.5455&0.5838&0.8463\\
GraphMMP (Ours) &\textcolor{red}{0.8535} &\textcolor{red}{0.7368} &\textcolor{red}{0.6796} &\textcolor{red}{0.7129}&\textcolor{red}{0.8649} \\ \hline
\end{tabular}
\label{comparison2}
\end{table}

\begin{figure}
\includegraphics[width=\textwidth]{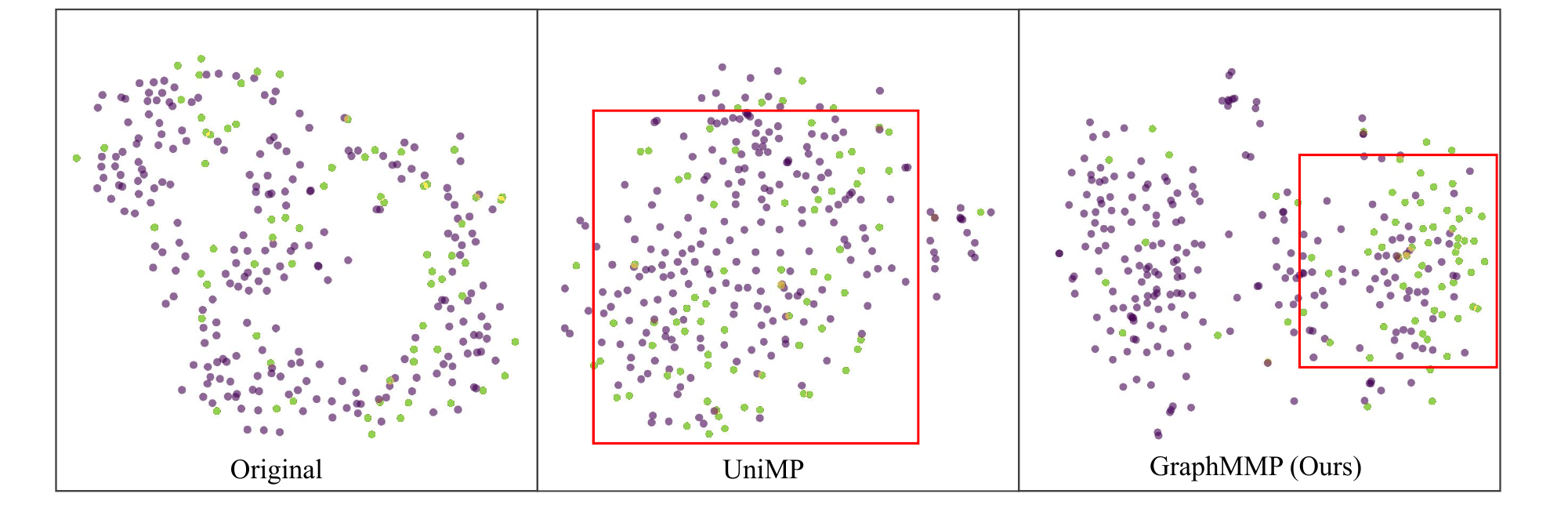}
\caption{t-SNE visualization of the best two methods on METABRIC. \textcolor{purple}{Purple} and \textcolor{green}{green} dots represent individual samples with different labels, respectively. The \textcolor{red}{red} box represents the approximate distribution of \textcolor{green}{green} samples.} \label{tsne}
\end{figure}

\textbf{Liver Prognosis dataset results.} Table~\ref{comparison} summarizes the performance of various methods on the liver prognosis task under different multimodal data configurations. The results demonstrate that GraphMMP consistently outperforms the compared methods. In scenarios where imaging data is unavailable, GraphMMP outperforms the leading Graph Neural Network models, GraphConv and UniMP. It demonstrates enhancements of 1.68\% in ACC, 0.0075 in F1-score, and 0.0176 in AUC, highlighting its effectiveness in handling non-visual data. Furthermore, when employing a full-modality approach, GraphMMP continues to excel by surpassing the second-best method, TMI-CLNet. It achieves improvements of 2.02\% in ACC, 0.0301 in F1-score, and 0.0188 in AUC. These results emphasize GraphMMP's superior performance in multimodal prognosis tasks, showcasing its ability to integrate and analyze diverse data types for more accurate predictions. These results highlight the robustness and superiority of GraphMMP in medical prognosis tasks, particularly within the GNN paradigm.

\textbf{METABRIC dataset results.} Table~\ref{comparison2} displays the comparative results of GraphMMP against other graph-based methods on the METABRIC dataset. GraphMMP outperforms the best baseline by 1.66\% in ACC, underscoring its advanced performance in medical prognosis tasks. This external validation on the METABRIC dataset also emphasizes the robustness and generalization ability of our method in multimodal medical prognosis.

\textbf{t-SNE visualization.} We employed t-SNE, with a perplexity of 30 and 300 iterations, to visualize and compare the input features and hidden representations of the two top-performing models. This visualization was conducted using a sample of 300 nodes. Fig.~\ref{tsne} shows distinct patterns: the left panel indicates significant class overlap in the original features, as seen in the purple and yellow dots. In contrast, UniMP's hidden layer features lack clear separability. GraphMMP, however, effectively clusters node features with different labels (highlighted by the red box), showing its ability to transform and refine input features into a more discriminative representation, capturing the underlying structure of the data.

\noindent\textbf{Ablation study.} Table~\ref{ablation} presents the results of ablation experiments to evaluate the importance of key components in the GraphMMP pipeline. These include the MI-based edge weights generation in Stage 1 and the inclusion of the MGF module in Stage 2. For comparison purposes, when MI is not applied as the edge weight, all edge weights are assigned a uniform default value of 1. When MI is employed for edge weight generation, GraphMMP shows ACC improvements by 4.06\% and 2.27\% on the liver prognosis and the METABRIC datasets, respectively, highlighting the benefit of incorporating implicit inter-modal relationships. Furthermore, adding the MGF module leads to additional ACC gains of 6.76\% and 3.03\% on the respective datasets. These results underscore the importance of global modeling capabilities in GNN layers, enhancing the model's representational power. Overall, the findings confirm the significant contributions of both components to GraphMMP's improved performance.

\begin{table}[t]
\centering
\caption{Ablation study of GraphMMP on its key components. The best and second-best results are shown in \textcolor{red}{red} and \textcolor{blue}{blue}, respectively.}
\begin{tabular}{ccccccc}
\hline
Dataset & Method & ACC & Precision & Recall & F1-score & AUC \\
\hline
\multirow{3}{*}{Liver} & GraphMMP w/o MI &\textcolor{blue}{0.8108} & \textcolor{blue}{0.8182} & \textcolor{blue}{0.6923} &\textcolor{blue}{0.7586}&\textcolor{blue}{0.7976}\\
& GraphMMP w/o MGF &0.7838 & 0.7501 & 0.6429 &0.7333&0.7500 \\
& GraphMMP (Ours) &\textcolor{red}{0.8514} & \textcolor{red}{0.8462} & \textcolor{red}{0.7857}&\textcolor{red}{0.8106}&\textcolor{red}{0.8411}\\
\hline
\multirow{3}{*}{METABRIC} & GraphMMP w/o MI &\textcolor{blue}{0.8308} & \textcolor{blue}{0.6957} & \textcolor{blue}{0.6214} &\textcolor{blue}{0.6564}&\textcolor{blue}{0.8556} \\
& GraphMMP w/o MGF &0.8232 & 0.6835 & 0.5455 &0.6067&0.8434 \\
& GraphMMP (Ours) &\textcolor{red}{0.8535} &\textcolor{red}{0.7368} &\textcolor{red}{0.6796} &\textcolor{red}{0.7129}&\textcolor{red}{0.8649} \\
\hline
\end{tabular}
\label{ablation}
\end{table}
\section{Conclusion}

In this study, we propose GraphMMP, a two-stage multimodal prognosis model based on GNNs. GraphMMP integrates a MI-based feature graph construction mechanism to capture hidden inter-modal relationships and a Mamba-based global fusion module to enhance global perception capabilities. Experimental results demonstrate superior performance in classification accuracy and robustness across multiple datasets, highlighting its effectiveness for multimodal medical prognosis tasks. Future work will expand the application of multimodal GNNs to a broader range of datasets to further enhance the generalizability and robustness of the approach in multimodal medical data analysis.

\begin{credits}
\subsubsection{\ackname} This study was funded by  the Open Project Program of the State Key Laboratory of CAD\&CG (No. A2410), Zhejiang University, Zhejiang Provincial Natural Science Foundation of China (No. LY21F020017), National Natural Science Foundation of China (No. 61702146, 62076084, U22A2033, U20A20386), Guangdong Basic and Applied Basic Research Foundation (No. 2025A1515011617, 2022A1515110570).

\subsubsection{\discintname}
The authors have no competing interests to declare that are
relevant to the content of this article.
\end{credits}

\bibliographystyle{splncs04}
\bibliography{1}

\begin{thebibliography}{10}
\providecommand{\url}[1]{\texttt{#1}}
\providecommand{\urlprefix}{URL }
\providecommand{\doi}[1]{https://doi.org/#1}

\bibitem{Behrad2022AnOO}
Behrad, F., Abadeh, M.S.: An overview of deep learning methods for multimodal medical data mining. Expert Systems with Applications  \textbf{200},  117006 (2022)

\bibitem{Breiman2001RandomF}
Breiman, L.: Random forests. Machine Learning  \textbf{45},  5--32 (2001)

\bibitem{Chandak2022BuildingAK}
Chandak, P., Huang, K., Zitnik, M.: Building a knowledge graph to enable precision medicine. Scientific Data  \textbf{10} (2022)

\bibitem{choi2021fully}
Choi, Y.S., Bae, S., Chang, J.H., Kang, S.G., Kim, S.H., Kim, J., Rim, T.H., Choi, S.H., Jain, R., Lee, S.K.: Fully automated hybrid approach to predict the idh mutation status of gliomas via deep learning and radiomics. Neuro-oncology  \textbf{23}(2),  304--313 (2021)

\bibitem{Cortes1995SupportVectorN}
Cortes, C., Vapnik, V.N.: Support-vector networks. Machine Learning  \textbf{20},  273--297 (1995)

\bibitem{cui2022survival}
Cui, C., Liu, H., Liu, Q., Deng, R., Asad, Z., Wang, Y., Zhao, S., Yang, H., Landman, B.A., Huo, Y.: Survival prediction of brain cancer with incomplete radiology, pathology, genomic, and demographic data. In: International Conference on Medical Image Computing and Computer-Assisted Intervention. pp. 626--635. Springer (2022)

\bibitem{curtis2012genomic}
Curtis, C., Shah, S.P., Chin, S.F., Turashvili, G., Rueda, O.M., Dunning, M.J., Speed, D., Lynch, A.G., Samarajiwa, S., Yuan, Y., et~al.: The genomic and transcriptomic architecture of 2,000 breast tumours reveals novel subgroups. Nature  \textbf{486}(7403),  346--352 (2012)

\bibitem{dosovitskiy2020image}
Dosovitskiy, A., Beyer, L., Kolesnikov, A., Weissenborn, D., Zhai, X., Unterthiner, T., Dehghani, M., Minderer, M., Heigold, G., Gelly, S., et~al.: An image is worth 16x16 words: Transformers for image recognition at scale. In: International Conference on Learning Representations (2020)

\bibitem{gu2023mamba}
Gu, A., Dao, T.: Mamba: Linear-time sequence modeling with selective state spaces. arXiv  (2023)

\bibitem{he2016deep}
He, K., Zhang, X., Ren, S., Sun, J.: Deep residual learning for image recognition. In: Proceedings of the IEEE/CVF Conference on Computer Vision and Pattern Recognition. pp. 770--778 (2016)

\bibitem{Hu2020HeterogeneousGT}
Hu, Z., Dong, Y., Wang, K., Sun, Y.: Heterogeneous graph transformer. Proceedings of The Web Conference 2020  (2020)

\bibitem{Kipf2016SemiSupervisedCW}
Kipf, T., Welling, M.: Semi-supervised classification with graph convolutional networks. ArXiv  \textbf{abs/1609.02907} (2016)

\bibitem{Li2024CGMegaEG}
Li, H., Han, Z., Sun, Y., Wang, F., Hu, P., Gao, Y., Bai, X., Peng, S., Ren, C., Xu, X., Liu, Z., Chen, H., Yang, Y., Bo, X.: Cgmega: explainable graph neural network framework with attention mechanisms for cancer gene module dissection. Nature Communications  \textbf{15} (2024)

\bibitem{li2022hfbsurv}
Li, R., Wu, X., Li, A., Wang, M.: Hfbsurv: hierarchical multimodal fusion with factorized bilinear models for cancer survival prediction. Bioinformatics  \textbf{38}(9),  2587--2594 (2022)

\bibitem{li2021braingnn}
Li, X., Zhou, Y., Dvornek, N., Zhang, M., Gao, S., Zhuang, J., Scheinost, D., Staib, L.H., Ventola, P., Duncan, J.S.: Braingnn: Interpretable brain graph neural network for fmri analysis. Medical Image Analysis  \textbf{74},  102233 (2021)

\bibitem{Morris2018WeisfeilerAL}
Morris, C., Ritzert, M., Fey, M., Hamilton, W.L., Lenssen, J.E., Rattan, G., Grohe, M.: Weisfeiler and leman go neural: Higher-order graph neural networks. In: AAAI Conference on Artificial Intelligence (2018)

\bibitem{Palmal2023BreastCS}
Palmal, S., Arya, N., Saha, S., Tripathy, S.: Breast cancer survival prognosis using the graph convolutional network with choquet fuzzy integral. Scientific Reports  \textbf{13} (2023)

\bibitem{shi2021masked}
Shi, Y., Huang, Z., Feng, S., Zhong, H., Wang, W., Sun, Y.: Masked label prediction: Unified message passing model for semi-supervised classification. In: Proceedings of the Thirtieth International Joint Conference on Artificial Intelligence. pp. 1548--1554. International Joint Conferences on Artificial Intelligence Organization (2021)

\bibitem{Tanvir2024MOGATAM}
Tanvir, R.B., Islam, M.M., Sobhan, M., Luo, D., Mondal, A.M.: Mogat: A multi-omics integration framework using graph attention networks for cancer subtype prediction. International Journal of Molecular Sciences  \textbf{25} (2024)

\bibitem{Topol2023AsAI}
Topol, E.J.: As artificial intelligence goes multimodal, medical applications multiply. Science  \textbf{381 6663},  adk6139 (2023)

\bibitem{Velickovic2023EverythingIC}
Velickovic, P.: Everything is connected: Graph neural networks. Current Opinion in Structural Biology  \textbf{79},  102538 (2023)

\bibitem{wu2025tmi}
Wu, L., Shan, X., Ge, R., Liang, R., Zhang, C., Li, Y., Elazab, A., Luo, H., Liu, Y., Wang, C.: Tmi-clnet: Triple-modal interaction network for chronic liver disease prognosis from imaging, clinical, and radiomic data fusion. arXiv preprint arXiv:2502.00695  (2025)

\end{thebibliography}

\end{document}